\documentclass[lettersize,journal]{IEEEtran}
\usepackage{amsmath,amsfonts}
\usepackage{algorithmic}
\usepackage{algorithm}
\usepackage{array}
\usepackage[caption=false,font=normalsize,labelfont=sf,textfont=sf]{subfig}
\usepackage[pagebackref=false,breaklinks=true,letterpaper=true,colorlinks,bookmarks=false,citecolor=blue,linkcolor=blue]{hyperref}

\usepackage{textcomp}
\usepackage{stfloats}
\usepackage{url}
\usepackage{verbatim}
\usepackage{graphicx}
\usepackage{cite}
\usepackage{microtype}
\usepackage{booktabs}
\usepackage[table]{xcolor}
\usepackage{arydshln}

\usepackage{mathptmx}
\usepackage{amssymb}
\usepackage{mathrsfs}
\usepackage{mathtools}
\usepackage{hyperref}

\usepackage{hyperref}

\usepackage{float}
\usepackage{colortbl}
\usepackage{booktabs}
\usepackage{stfloats}
\usepackage{amsfonts}
\usepackage{makecell}

\begin{document}

\title{QR-CLIP: Introducing Explicit Knowledge for Location and Time Reasoning}

\author{Weimin Shi, Mingchen Zhuge, Dehong Gao, Zhong Zhou, \\Ming-Ming Cheng~\IEEEmembership{Senior Member,~IEEE}, Deng-Ping Fan~\IEEEmembership{Senior Member,~IEEE}
\thanks{The first two authors share equal contributions. Weimin Shi, Zhong Zhou are with State Key Laboratory of Virtual Reality Technology and System, Beihang University, Beijing, China.
Mingchen Zhuge is with AI Initiative, King Abdullah University of Science and Technology (KAUST), Saudi Arabia.
Dehong Gao is with Northwestern Polytechnical University, Xian, China.
Ming-Ming Cheng is with the Nankai University, Tianjin, China.
Deng-Ping Fan is with the
Computer Vision Lab (CVL), ETH Zurich, Zurich, Switzerland.
 Corresponding author: Zhong Zhou (zz@buaa.edu.cn).
}}

\markboth{Journal of \LaTeX\ Class Files,~Vol.~14, No.~8, August~2021}%
{Shell \MakeLowercase{\textit{et al.}}: A Sample Article Using IEEEtran.cls for IEEE Journals}


\maketitle

\begin{abstract}
Daily images can convey abstract meanings that require us to memorize and infer profound information. To encourage human-like reasoning, in this work, we teach machines to predict where and when the image was captured.
Inspired by Hutchins's theory of Distributed Cognition, we design a new model called \textbf{QR-CLIP}, which is composed of two components.
1) The \textbf{Quantity} Module expands cognitive resources by accumulating as much open-world knowledge as possible from the environment, thereby enhancing cognitive ability.
2) The \textbf{Relevance} Module integrates relevant information from various cognitive tools to create a comprehensive cognitive outcome.
Experiments have shown the effectiveness of our QR-CLIP, which outperforms the previous SOTA methods on each task, achieving an average relative improvement of  $\sim$10\% and $\sim$110\% in terms of location and time reasoning, respectively.
This study provides a technical foundation for location and time reasoning and suggests that effectively interacting with open-world knowledge is one of the solutions for the tasks. The source code is available at \href{https://github.com/Shi-Wm/QR-CLIP}{https://github.com/Shi-Wm/QR-CLIP}.
\end{abstract}


\section{Introduction}
\IEEEPARstart{M}{any} deep computer vision models possess outstanding perception abilities and can solve routine tasks by extracting basic visual contexts (\emph{i.e.}, color, texture, and objects), following the principle: ``what you see is what you get."
However, they cannot engage with a scene in the same insightful ways as humans can~\cite{crowder2012artificial, hassabis2017neuroscience, wirtz2018brave}. It seems difficult for them to think deeply based on their observations~\cite{salovey1990emotional, schmidhuber2015learning, harrer2019artificial}.

\begin{figure}[t!]
\begin{center}
\includegraphics[width=1.0\linewidth]{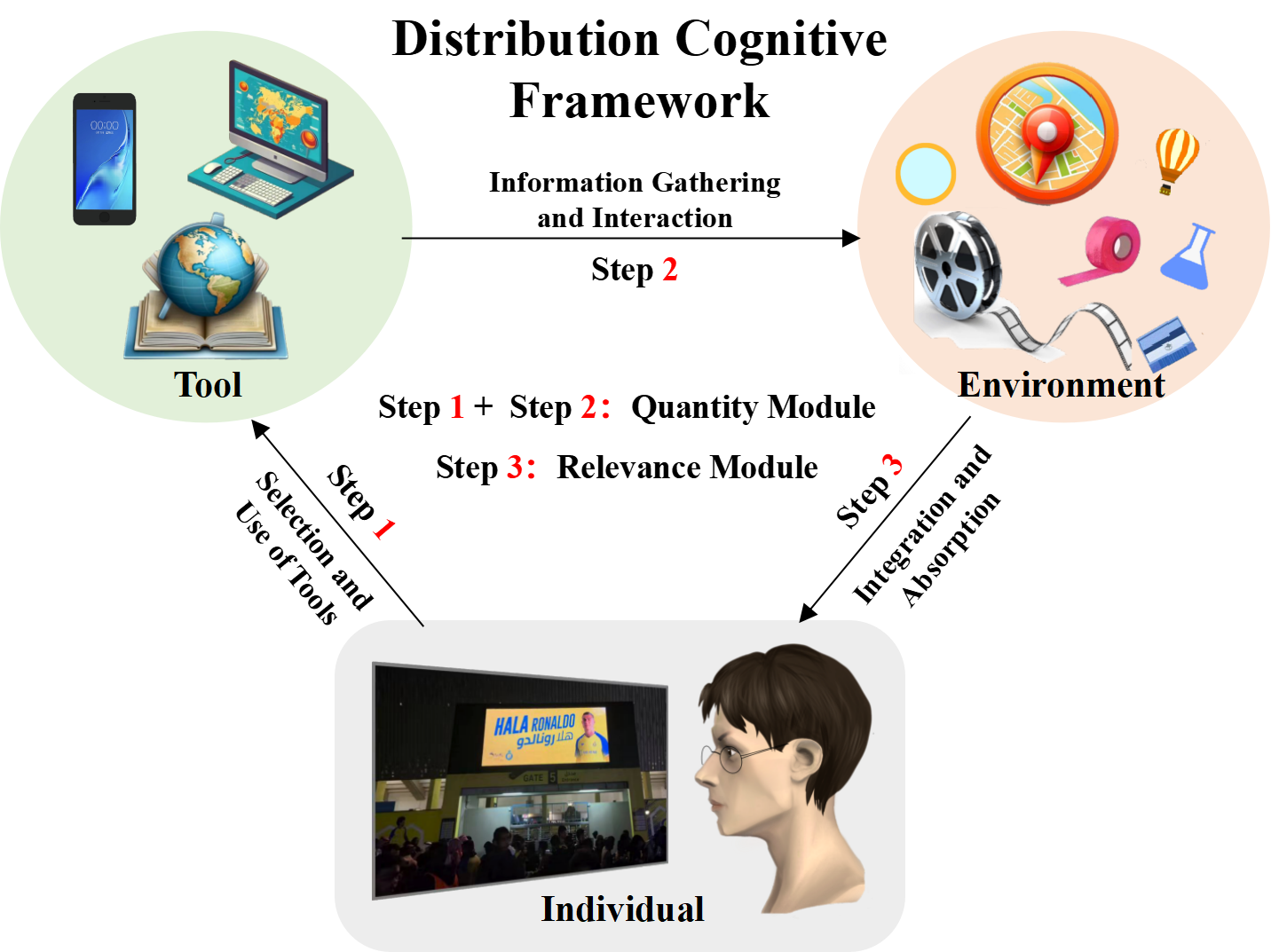}
\end{center}
\vspace{-15pt}
    \caption{
    The diagram of the principle of distributed cognition. The individual, positioned at the central locus, engages in interactive exchanges with the environment when presented with an image, utilizing diverse tools such as smartphones, books, and computers. Through this interactive process, the individual acquires a comprehensive and profound cognition of the image's information. The framework emphasizes the pivotal role of tool-mediated interactions with the environment in facilitating individual cognitive processes.
    }
\label{fig:DC_theory}
\end{figure}

\begin{figure*}[t!]
\begin{center}
\includegraphics[width=\linewidth]{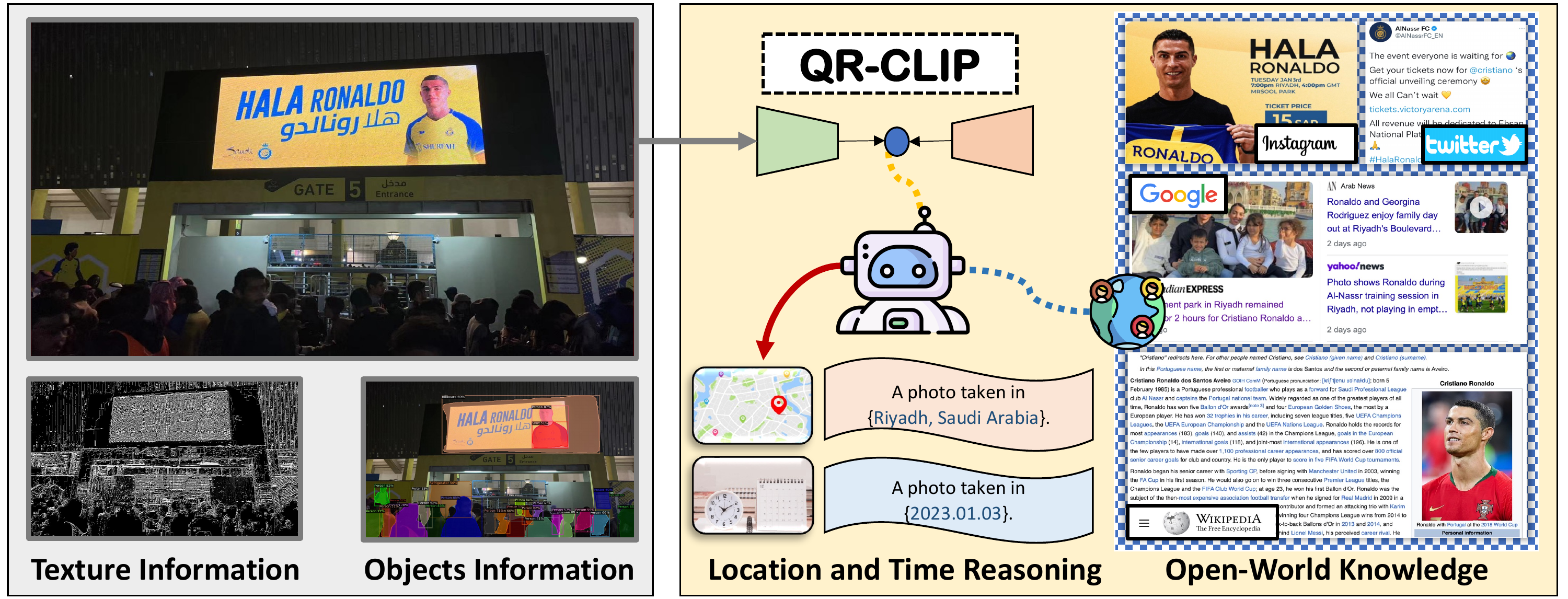}
\end{center}
\vspace{-15pt}
    \caption{
    Comparing traditional computer vision tasks (\emph{left}) with location and time reasoning (\emph{right}), it becomes clear that location and time reasoning requires more human experience and knowledge, (\emph{a.k.a.} open-world knowledge, rather than just simple image color, texture, and object information.
    }
\label{fig:pipeline}
\end{figure*}

In pursuit of the goal of enabling machines to emulate human-like depth of thought, this paper draws inspiration from the human cognitive process and is grounded in the distributed cognition theory~\cite{hutchins1991social, hutchins1995cognition, hutchins2000distributed} (as illustrated in Fig.~\ref{fig:DC_theory}). The framework established in this paper explores the model's ability to infer underlying information from images, with a focus on the reasoning of location and time information behind the images~\cite{fu2022there}.
The procedure can be summarized as follows: \emph{input an image, the model's goal is to guess where and when the image was taken.} This task is quite different from others (\emph{e.g.,} basic classification~\cite{wang2021n15news, yang2022coarse, sun2021supervised, pei2019effects, shoeybi2019megatron}, summarization~\cite{fabbri2019multi, jiang2022joint, ma2022multi}, or retrieval tasks~\cite{conforti2020stander, zuo2023fine, ma2022multi}), since it requires the model to delve deeper into the information and truly comprehend the event behind the images. 
As shown in Fig.~\ref{fig:pipeline}, the model must not only comprehend the image details, such as Cristiano Ronaldo and the language on the signs, but also search for text related to the image in the environment. By combining this information, the model can reason that the photo was taken at the unveiling ceremony of Cristiano Ronaldo joining Al-Hilal Riyadh.

These days, industries invest millions of dollars in training foundation models~\cite{reed2022generalist, schwartz2020green, strubell2019energy}. Advanced parallel techniques~\cite{rasley2020deepspeed, ott2019fairseq, huang2019gpipe} enable the model to scale up the number of parameters and data.
Some companies, such as OpenAI and DeepMind, have developed a range of models, including GPT-3~\cite{brown2020language}, CLIP~\cite{radford2021learning}, and ChatGPT~\cite{ouyang2022training}, among others~\cite{nichol2022point, radford2022robust, dhariwal2020jukebox, rae2021scaling, borgeaud2022improving, anil2023palm}. 
Most of these foundational models learn independently from large amounts of online data created by people in a self-supervised manner. 
In this paper, we delve into acquiring knowledge by models from the environment, which we refer to as ``open-world knowledge" (OWK). This terminology emphasizes that models access and utilize information resources in a dynamic and constantly evolving world. The OWK enables models to understand the environment comprehensively.
This motivates us to utilize CLIP~\cite{radford2021learning} as the foundational architecture for addressing the proposed task, owing to its proven efficacy in a broad spectrum of multimodal applications.

Compared to traditional image models such as ResNet and Swin~\cite{he2016deep,liu2021swin}, CLIP initially contains a certain amount of OWK. However, this knowledge is more encapsulated within the model and can only be used implicitly through in-context learning~\cite{min2022rethinking}, which prevents OWK from playing a larger role in our tasks.
To tackle this problem, we introduce a novel model, \textbf{QR-CLIP}, which can accurately determine an image's location and time-related meta information.
It is inspired by Hutchins' distributed cognition theory~\cite{hutchins1991social, hutchins1995cognition, hutchins2000distributed}. This theory reveals how human cognitive processes interact among individuals, tools, and the environment to extend and enhance cognitive abilities. Within the framework of distributed cognition, we have designed two modules: the \textbf{Quantity} Module and the \textbf{Relevance} Module. The \textbf{Quantity} Module facilitates individuals' accumulation of as much knowledge as possible from external resources and the environment using various tools. This module is critical in expanding cognitive resources and enhancing cognitive ability. On the other hand, the \textbf{Relevance} Module embodies the process of consolidating and selecting pertinent information provided by different cognitive tools to form an integrated cognitive outcome. This module enhances cognitive effectiveness while simultaneously reducing uncertainty and bias.

To better understand our \textbf{QR-CLIP} model, we now elaborate on the working principle of the \textbf{Quantity} Module.
This module employs two techniques to maximize the capture of environmental information. For traditional transformer-based models, such as BERT~\cite{devlin2018bert} and image transformer~\cite{dosovitskiy2020image}, a single \texttt{[CLS]} token is used to represent the input.
Initially, we developed additional \texttt{[CLS]} tokens as cognitive tools to simulate various human cognitive perspectives on the same image. Each cognitive tool processes and interprets input information uniquely, facilitating a more comprehensive understanding of a specific item when combined with knowledge from diverse viewpoints. In the context of distributed cognition theory, cognition surpasses the confines of an individual's brain and is realized through interactions with the environment. This concept further inspired us to utilize each \texttt{[CLS]}$_i$ token to retrieve valuable OWK, which aids in reasoning.
Motivated by current contrastive learning methods~\cite{zhang2022multi, chen2020simple, he2020momentum, xia2023graph}, we have designed local and global loss functions for fine-tuning the CLIP model, ensuring that our \textbf{QR-CLIP} is suitable for the intended tasks.

For the \textbf{Relevance} Module, we design a scoring mechanism that weighs the fusion of image and OWK embeddings. Just as in any environment, not all knowledge is correct. Therefore, our scoring mechanism serves as an error correction tool to assist the model in selecting the most valuable information. It adaptively balances different types of information and encourages the model to provide relevant information for location and time reasoning. 
Additionally, the scoring mechanism maintains a balance between visual and language knowledge, signifying that in circumstances where the quality of explicit OWK is less than satisfactory, our scoring mechanism can place increased emphasis on original image features, thereby markedly enhancing model robustness.

The experiments have shown the strong abilities of our \textbf{QR-CLIP} model. In terms of accuracy (or Rank@1), it achieved 19.51\%, which is an 18.55\% relative improvement compared to the previous SOTA on location reasoning. 
Moreover, for for time reasoning tasks, our model achieved an accuracy of 3.45\%, representing a significant relative improvement of 245\%.

\section{Related Work}
\subsection{Foundation Models}

The emergence of foundation models is a relatively recent phenomenon~\cite{bommasani2021opportunities, arora2022ask, zhou2023comprehensive}, fundamentally altering the game rules of AI communities. They are commonly trained on a massive amount of unlabeled data at scale (typically via self-supervised learning~\cite{zhuge2021kaleido, chen2022scaling, luo2022siman, sirotkin2022study, paul2022self, xia2022skating, ji2023masked, wang2023self}), making them adaptable to various downstream applications. 
Among the popular models, GPT-3~\cite{brown2020language} and PaLM~\cite{chowdhery2022palm} are renowned for their expansive understanding and generation of human language as large language models.
Models such as CLIP~\cite{radford2021learning} and Flamingo~\cite{alayrac2022flamingo} are vision-language models that show impressive capabilities in understanding and relating visual and textual information, bridging the gap between these two key data modalities. In the realm of text-to-image generation, models such as Dall-E~\cite{ramesh2021zero} and Stable Diffusion~\cite{rombach2022high} have set new benchmarks by exhibiting the capacity to generate intricate and contextually relevant images from textual descriptions.
GaTo~\cite{reed2022generalist}, classified as a generalist model, is renowned for its extensive knowledge base and adaptability across various tasks, showcasing the versatility of foundational models.
ChatGPT~\cite{ouyang2022training} as a human-like conversation agent shows versatility by understanding and responding to diverse topics and contexts, \emph{etc}.

This paper uses CLIP pre-trained with $400$ million image-text pairs as baseline architecture. It learns excellent OWK and multi-modal representation by learning with such a large-scale corpus, making it ideal as the basic solution. Based on it, we made \textbf{QR-CLIP} to fit the location and time reasoning tasks. By integrating \textbf{QR-CLIP}, we aim to further enhance the ability of foundation models to cognize and reason about abstract information behind images.

\subsection{Location and Time Reasoning}

Existing language models have achieved significant success in a wide range of tasks that require an understanding of language~\cite{devlin2018bert,lewis2019bart, wang2018glue}. Also, the vision models~\cite{he2016deep,dosovitskiy2020image,liu2021swin} can predict the correct class label of an image from thousands of options. However, they still face challenges in performing various reasoning tasks, such as discerning the abstract meanings (\emph{e.g.,} time, location, event) of images~\cite{yang2020learning,tahmasebzadeh2021geowine}, conducting mathematical calculations~\cite{lewkowycz2022solving}, and making scientific deductions~\cite{degrave2022magnetic}. 
However, humans excel at these tasks due to the superior computational power of real brains compared to artificial neural networks, as well as their ability to actively learn and apply abstract reasoning~\cite{schmidhuber2015learning}.

In the realm of location and time reasoning, numerous studies have been conducted to extract spatial and temporal information from text. These efforts include predicting user locations from social media text~\cite{chen2019region}, extracting time information from diverse texts~\cite{zhou2020temporal, han2020econet}, and inferring spatiotemporal quantities from news articles~\cite{ning2022meta}. In our research, we focus on location and time reasoning~\cite{fu2022there} from images. This requires a model to think and reason beyond the actual content of an image. Compared to other reasoning tasks, this task differs in---there are often insufficient visual cues to make reasonings, necessitating the use of auxiliary knowledge.

\section{Approach}

\subsection{Preliminary}
\textbf{Task Background.} 
The Current AI methods are relatively weak in cognizing and reasoning the abstract information concealed within an image. 
The goal of this paper is to let the model reason the location and time based on image input~\cite{fu2022there}: given an image $I$, we need the model ($\text{M}(I)$) to predict the location ($\text{Pred}_{l}$) and time ($\text{Pred}_{t}$).

\textbf{Distributed Cognition Theory.} 
The theory of distributed cognition suggests cognition as a process distributed among individuals, tools and the environment~\cite{hutchins1991social, hutchins1995cognition, hutchins2000distributed}. This is crucial for understanding human cognition and reasoning, emphasizing  the necessity of the interactions among these components.

As shown in Fig.~\ref{fig:DC_theory}, individuals play a central role in the cognitive process as the primary site for information processing and reasoning. Concurrently, everyday tools such as mobile phones, computers, and books play a pivotal role in enhancing cognitive abilities by assisting in the storage, processing, and retrieval of environmental knowledge, thereby expanding cognitive capabilities. Additionally, the environment, including the physical environment and social context, is an integral part of distributed cognitive processes and also affects the cognitive function of individuals.

Within the framework of distributed cognition theory, we introduce the \textbf{QR-CLIP} method, which aims to mimic the human process of distributed cognition to deduce the underlying information within images.

\begin{figure*}[t!]
\begin{center}
\includegraphics[width=1.0\linewidth]{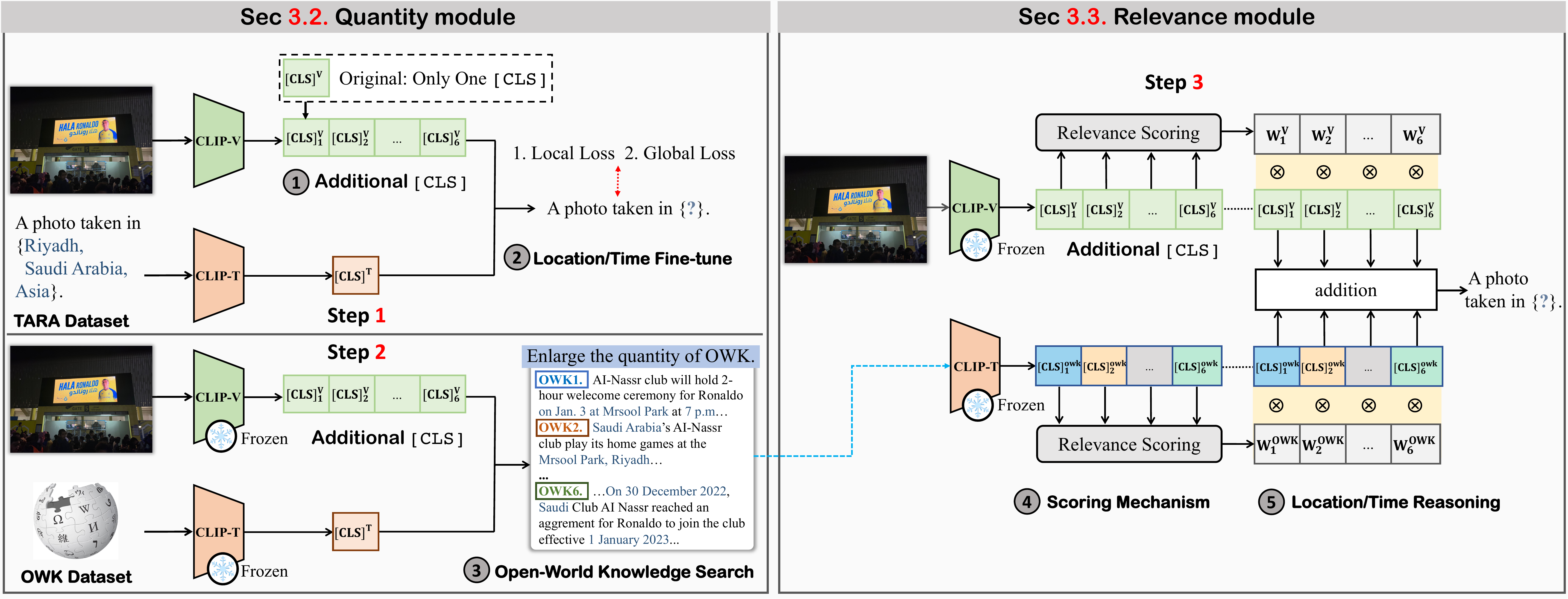}
\end{center}
\vspace{-15pt}
    \caption{
    The \textbf{QR-CLIP} pipeline consists of two Modules: the Quantity Module (Sec~\ref{sec:qm}) and the Relevance Module (Sec~\ref{sec:rm}). 
    Each step described below corresponds to Step \textcolor{red}1, Step \textcolor{red}2, and Step \textcolor{red}3 in Fig.~\ref{fig:DC_theory}.
    In Step \textcolor{red}1, we add additional \texttt{[CLS]} tokens to simulate the use of different cognitive tools by individuals. We then design local and global loss functions to guide location/time fine-tuning.
    Then, we freeze the fine-tuned CLIP-V and CLIP-T models, and utilize them to search for OWK from our OWK dataset (Sec.~\ref{sec:data}).
    In the Relevance Module, we use a scoring mechanism to weigh the most valuable information from CLIP-T and CLIP-V.
    After multiplying the scoring weights for vision and language separately, we add them for the final similarity calculation. 
    }
\label{fig:model}
\end{figure*}

\textbf{Our Pipeline.} 
The \textbf{QR-CLIP} is based on the CLIP \cite{radford2021learning}, which was pre-trained on a corpus of 400 million internet data. As shown in Fig.~\ref{fig:model}, it consists of two modules: the Quantity Module and the Relevance Module.

1) The Quantity Module aims to expand the cognitive resources of the model. To achieve this, we propose utilizing the additional \texttt{[CLS]} method, which generates multiple distinct \texttt{[CLS]}$_i$ as cognitive tools within the model. These tools approach the search for OWK from various perspectives.

2) The Relevance Module incorporates a scoring mechanism to assess the retrieved OWK and image features, guaranteeing the selection of the most pertinent information. These chosen features are then aggregated for final reasoning.

\subsection{Quantity Module}\label{sec:qm}

\textbf{Additional \texttt{[CLS]}.} We utilize CLIP~\cite{radford2021learning} as our foundational architecture. In vanilla CLIP, $\Hat{\texttt{[CLS]}}$ is used to differentiate the input token that represents the entire input features, which is a common practice in other transformer models~\cite{devlin2018bert,dosovitskiy2020image}. 
To simplify the illustration and better represent our model in the subsequent sections, we define $[\texttt{CLS}]^{v}_{i}$ and $[\texttt{CLS}]^{t}$ as:

\begin{equation}
[\texttt{CLS}]^{v}_{i}\leftarrow\text{Enc}_v(\Hat{[\texttt{CLS}]_i^{v}})  \text{ and } [\texttt{CLS}]^{t}\leftarrow\text{Enc}_t(\Hat{[\texttt{CLS}]^{t}}),
\end{equation}
where $\text{Enc}_v$ represents the image encoder, and $\text{Enc}_t$ represents the text encoder, both of which are components of CLIP.
It is worth noting that when referring to $[\texttt{CLS}]$ ($[\texttt{CLS}]^{v}_{i}$ and $[\texttt{CLS}]^{t}$) by default, they represent CLIP's output embeddings rather than the input token ($\Hat{\texttt{[CLS]}}$), as shown in Fig.~\ref{fig:model}.

However, the output embedding \texttt{[CLS]} is inadequate in representing an image as a single embedding provides limited location and time reasoning cues.
Therefore, we propose to expand the image representations.
It is evident that in real life, 
individuals can achieve a more comprehensive and accurate understanding of images by integrating the information and functionality of various tools.
In this vein, we propose a simple yet effective method. In our technical implementation, we introduce additional $\texttt{[CLS]}_i^{v}$ tokens, where $i$ represents the count of \texttt{[CLS]} tokens in a given image, ranging from 1 to $n$. This strategy replaces the conventional single \texttt{[CLS]} representation method.

Through the observation of ablations, we finally use six different $\Hat{\texttt{[CLS]}_i^{v}}$ at the beginning of the image patch token embeddings ($\Hat{I}={\Hat{I}^{patch}_{1}, ..., \Hat{I}^{patch}_{7}}$), like $(\Hat{[\texttt{CLS}]_1^{v}}...\Hat{[\texttt{CLS}]_6^{v}} \ \Hat{I})$. After passing through the encoder $\text{Enc}_v$, we get a list of embeddings $([\texttt{CLS}]_1^{v}...[\texttt{CLS}]_6^{v} \ I)$. 
Using this design, 
each $\texttt{[CLS]}_i^{v}$ token is treated as a separate cognitive tool, simulating the framework of distributed cognition. This approach enables the pre-trained model to incorporate multiple perspectives, enhancing the richness of the captured OWK.

Since the text contains explicit semantic information and most language inputs convey clear messages, we only utilize the original $[\texttt{CLS}]^{t}$ at the beginning of the text token embedding (T), denoted as $([\texttt{CLS}]^{t}\ T)$.  Therefore, we search for corresponding information in the image from the CLIP model:

\begin{equation}
([\texttt{CLS}]^{t}) \cdot ([\texttt{CLS}]_i^{v}), \label{search}
\end{equation}
here, the $\cdot$ denotes the inner product operation. In the fine-tuning or OWK search process, each $[\texttt{CLS}]_i^v$ from $\text{Enc}_v$ calculates its similarity with the $[\texttt{CLS}]^t$ of the candidate information.

\textbf{Location/Time Fine-tune.}                 
First, we individually initialize and position-encode each $[\texttt{CLS}]_i^v$ to increase the distance between them.
We fine-tune CLIP with local and global losses~\cite{he2020momentum,zhang2022multi} to ensure that each $[\texttt{CLS}]_i^v$ is aligned with the linguistic features of location and time  $[\texttt{CLS}]^t$. 

The local loss is utilized to construct different $[\texttt{CLS}]_i^v$ to comprehensively preserve visual features from diverse images' perspectives. This loss function consists of multi-view contrastive learning (MVC) and multi-view regularization (MVR).
The correspondence between each $[\texttt{CLS}]_i^v$ and $[\texttt{CLS}]^t$ is achieved through the multi-view contrastive learning loss:
\begin{equation}
\begin{array}{l}
L_{MVC}=-\text{log}\dfrac{e^{f(q_{v}^i, \;k_{t+})}}{e^{f(q_{v}^i, \;k_{t+})}+e^{f(q_{v}^i, \;k_{t-})}}, \label{MVC_loss}
\end{array}
\end{equation}
here, $q_{v}^i$ denote the query image embedding ($[\texttt{CLS}]_i^v$), while $k_{t+}$ and $k_{t-}$  represent the positive and negative key text embeddings, in a batch of $[\texttt{CLS}]^t$.  The correlation score is calculated using the inner product function $f(x, y)$.

However, since multiple $[\texttt{CLS}]_i^v$ correspond to one $[\texttt{CLS}]^t$, the model tends to cluster $[\texttt{CLS}]_i^v$ together in the loss calculation according to Eq.~\ref{MVC_loss}. This clustering makes it difficult to distinguish information from different angles.
To overcome this issue, we use a regularization loss function to separate the distance between each $[\texttt{CLS}]_i^v$, promoting the independent learning and representation of diverse perspectives information in the image by the model:
\begin{equation}
\begin{array}{l}
L_{MVR}=\frac{2}{n(n-1)}\sum_{i=1}^{n-1}\sum_{j=i+1}^{n}\frac{f\left(q_{v}^i, \;q_{v}^j\right)}{||q_{v}^i||\cdot||q_{v}^j||}, \label{MVR_loss}
\end{array}
\end{equation}
where $n$ represents the number of $[\texttt{CLS}]_i^v$ vectors, $i$ ranges from 1 to $n-1$, and $j$ range from $i+1$ to $n$. 
This implies that the calculation of $L_{MVR}$ takes into account all possible pairs of $[\texttt{CLS}]_i^v$ vectors.
The numerator of the formula uses the inner product function, $f(q{v}^i, \;q_{v}^j)$, to gauge the similarity between each pair of $[\texttt{CLS}]_i^v$ vectors. 
The denominator normalizes the distance between the vectors by the product of their magnitudes $||q{v}^i||\cdot||q_{v}^j||$, which encourages the model to learn discriminative features that are independent of vector length.

During the optimization phase, the model strives to increase the distances between pairs of $[\texttt{CLS}]_i^v$, thereby enhancing its ability to discern image features from diverse perspectives. However, each $[\texttt{CLS}]_i^v$ vector faces distinct learning challenges, resulting in uneven rates of training progress. To counteract this, we use a dynamically balanced learning strategy: 
\begin{equation}
\begin{array}{l}
w^{i}=\mathrm{softmax}(1-acc^{i}), \label{dynamically balanced}
\end{array}
\end{equation} 
the variable $w^{i}$ denotes the dynamic learning weight for each $[\texttt{CLS}]_i^v$. 
This weight is calculated by applying a softmax function over the $(1-acc^{i})$, where $acc^{i}$ represents the accuracy of the model about each $[\texttt{CLS}]_i^v$. 
In essence, this approach dynamically adjusts the learning priorities, offering more attention to  instances of $[\texttt{CLS}]_i^v$ that  exhibit slower progress or present greater learning challenges.

Based on the above-mentioned discussions, the local loss can be defined as:
\begin{equation}
\begin{array}{l}
L_{local} =  \sum_{i=1}^{n} w^i L_{MVC} + \lambda L_{MVR},
\end{array}
\end{equation}
 it aims to minimize the distance between each $[\texttt{CLS}]_i^v$ and its corresponding sentence embedding ($[\texttt{CLS}]_t$), while simultaneously maximizing the distance between different $[\texttt{CLS}]_i^v$.

Then, the global loss serves to further constrain the correspondence between image features and location/time features. The calculation for this constraint is as follows:

\begin{equation}
\begin{array}{l}
L_{global}=-\text{log}\dfrac{e^{f_{mean}(q_v,\;k_{t+})}}{e^{f_{mean}(q_v, \;k_{t+})}+e^{f_{mean}(q_v, \;k_{t-})}}, \label{global_loss}
\end{array}
\end{equation}
here, we have the function $f_{mean}(q_v, k_t) =\frac{1}{n} \sum_{i=1}^{n} f(q_v^i, k_t) $. 
Next, the global loss function aims to consolidate the learning of global feature correspondence by integrating the alignment between the image and the text. This is represented by the mean correlation score across all perspectives. This approach fosters a more comprehensive understanding by merging insights from various viewpoints.

The entire training loss is defined as a linear combination of two losses as $L_{total} =  L_{local}  + L_{global}$. 
This composite loss function not only encourages the model to learn robust correspondences between diverse visual perspectives and the text but also enables it to capture the overall alignment of the image features with respect to location and time attributes mentioned in the text.

\textbf{Open-World Knowledge (OWK) Search.}
After fine-tuning, each $[\texttt{CLS}]_i^v$ outputted by CLIP-V is capable of representing image location and time information from various perspectives. 
We use these different representations to retrieve more valuable open-world knowledge from the OWK dataset (Sec~\ref{sec:data}), facilitating interaction with the environment.

Given an image $I$ and its corresponding OWK ($O={T_1^{owk}, T_2^{owk}, ..., T_k^{owk}}, k=122,408$), the search process follows Eq.~\ref{search}: each $[\texttt{CLS}]_i^v$ calculates the similarity with 122,408 candidate Wikipedia corpus (OWK).
Here, we select the  Wikipedia candidate with the highest similarity for each $[\texttt{CLS}]_i^v$, yielding a total of 6 OWK.
After that, the Quantity Module (sec~\ref{sec:qm}) completed its task of gathering a list of highly-related OWK items. This list will serve as input for the next module, the Relevance Module (sec~\ref{sec:rm}).

\subsection{Relevance Module}\label{sec:rm}
\textbf{Scoring Mechanism.} 
Within the framework of distributed cognitive principles, our approach, \textbf{QR-CLIP}, employs various cognitive tools, including multiple distinct $[\texttt{CLS}]_i^v$, to acquire open-world knowledge. The effectiveness of these tools varies based on the image and its corresponding embeddings of open-world knowledge. Thus, it is crucial to dynamically evaluate the significance of different features. Motivated by this, we propose a scoring mechanism to further emphasize and highlight relevant features.

We adopt two-layers \texttt{MLP} ($\text{MLP}_{2-layer}$) as our relevance scoring component and find it to be beneficial:

\begin{equation}
\begin{array}{l}
{W}^x = \text{MLP}_{2-layer}([\texttt{CLS}]^x_{i}),
\end{array}
\end{equation}
here, $[\texttt{CLS}]^x_{i}$ is the input embedding, and  ${W}^x$ is the calculated weight. We optimize the model using contrastive learning. To simplify implementation, we directly adopt the loss functions from the first step of the \textbf{Quantity} Module~(Sec~\ref{sec:qm}). 
In this case, we maintain the CLIP-T and CLIP-V frozen, and solely update the parameters of the relevance scoring component.

In the local loss, the information of two features is integrated to jointly optimize the scoring mechanism:
\begin{equation}
\begin{array}{l}
f(q_i, \;k_+) = (W_i^{owk} \times [\texttt{CLS}]_i^{owk} + W_i^{v} \times [\texttt{CLS}]_i^{v}) \cdot F^{gt},
\end{array}
\end{equation}
$W_i^{owk}$ and  $W_i^{v}$ are the weights of the $[\texttt{CLS}]_i^{owk}$ and  $[\texttt{CLS}]_i^{v}$;  $q_i$ in this place is the sum of weight vision-language features $W_i^{owk} \times [\texttt{CLS}]_i^{owk} + W_i^{v} \times [\texttt{CLS}]_i^{v}$. Additionally, $F^{gt}$ denotes the ground-truth features generated by $F^{GT}=\text{Enc}_t(GT)$.

We employ the fused features $F^{fused} = \sum_1^6 (W_i^{owk} \times [\texttt{CLS}]_i^{owk} + W_i^{v} \times [\texttt{CLS}]_i^{v})$ as our final features for reasoning the location and time. Next, we calculate the similarity between $F^{fused}$ and the embeddings of candidate locations/times to complete the reasoning.

We argue that by utilizing the CLIP (which is pre-trained $400$M open-world corpus) and subsequently fine-tuning it by adding additional \texttt{[CLS]} with location-and-time-specific data, the model can reason about meta information more effectively. 
\textbf{QR-CLIP} can then enhance its performance by retrieving valuable OWK and utilizing it as auxiliary cues.
Finally, the model balances vision and language embeddings, and by incorporating them into reasoning, the model attains a more effective performance.
The process is related to distributed cognition theory.
It is important to note that the procedure of information spreading is mimicked by~\cite{zhang2006distributed}: 
individuals who utilize multiple tools can generate diverse viewpoints and attitudes towards the same object (Sec \ref{sec:qm}). However, effectively combining these viewpoints can promote more comprehensive cognition (Sec \ref{sec:rm}).

\section{Experiments}

\subsection{Training Settings} \label{sec:data}
\textbf{Dataset.} We used two datasets: TARA dataset~\cite{fu2022there} and OWK dataset.
\textbf{TARA dataset} includes 15,429 samples, each consisting of a news picture and its corresponding location and time description. Following the original setup, we train \textbf{QR-CLIP} on a training set containing 12,306 instances and evaluate our method using a test set containing 1,644 instances.
The \textbf{OWK dataset} is derived from the WIT dataset~\cite{srinivasan2021wit}. Due to the limited computational resources, we have selected  122,408 texts from the 37.5 million entity-rich image-text examples in English Wikipedia that correspond to the specific countries and years as our OWK.

\textbf{Evaluation Metrics.}
For a fair comparison, we first follow the same evaluation metrics as outlined in the TARA benchmark~\cite{fu2022there}: Accuracy and Example-F1. Accuracy is calculated by comparing the predicted results with the entire labels. Example-F1 is calculated by comparing reasoning results with hierarchical labels:

\begin{equation}
\begin{array}{l}
\operatorname{Example-F1}=\dfrac{1}{N}\sum\limits_{i=1}^N\dfrac{2\left|\text{GT}_{i}\cap\text{Pred}_{i}\right|}{\left|\text{GT}_{i}\right|+{\left|\text{Pred}_{i}\right|}},
\end{array}
\end{equation}
where $\text{GT}_{i}$ represents the hierarchical label, and $\text{Pred}_{i}$ represents the hierarchical reason. If the entire label is \{\emph{`Zurich, Switzerland, Europe'}\}, the progressive hierarchical labels consist of three combinations of true labels: \{\emph{`Zurich, Switzerland, Europe'\}}, \{\emph{`Switzerland, Europe'}\} and  \{\emph{`Europe'}\}.  In addition, the proposed method's performance is evaluated using Rank@5 and F1-Score.

\textbf{Implementation Details.}
\textbf{QR-CLIP} is based on CLIP+VIT-B/32 model with an input size of $224 \times 224$. It is implemented on the PyTorch 1.10.1 platform with the Adam optimizer to update the neural network's weights and biases. 
The training batch size is $32$, and the initial learning rate is $1e-6$. Our model utilizes a pre-trained model and fine-tune process on an NVIDIA RTX 3090 GPU running CUDA 11.7.1.

\begin{table*}[!t] 
    \centering
    \caption{Summary of the performance for different baselines on the image location and time reason. The symbol $\dagger$ denotes that the original CLIP~\cite{radford2021learning} was fine-tuned. The term `\emph{AVG}' represents the average relative lift.}
    \label{tab:ModelSummary}
    \resizebox{\linewidth}{!}{
        \setlength\tabcolsep{10pt}
        \begin{tabular}{ll|c||c|c|c|c}
            \hline
            \cellcolor[HTML]{CBCEFB}ID   &\cellcolor[HTML]{CBCEFB} Method      & \cellcolor[HTML]{CBCEFB}Training Mode    & \cellcolor[HTML]{CBCEFB}Accuracy (Rank@1)     &\cellcolor[HTML]{CBCEFB} Rank@5      &\cellcolor[HTML]{CBCEFB} Example-F1   & \cellcolor[HTML]{CBCEFB}F1-Score  \\ \hline\hline
            \multicolumn{7}{c}{\cellcolor[HTML]{DAE8FC}Location Reasoning} \\ \hline
            1    & ResNet-50~\cite{he2016deep}  & Fine-tune  & 3.18\%         & 9.82\%          & 22.19\%     &  2.27\%     \\
            2    & Swin-T~\cite{liu2021swin}      & Fine-tune  & 6.70\%          & 17.07\%         & 33.56\%    & 5.02\%       \\
            3    & CLIP~\cite{radford2021learning}        & Zero-Shot   & 11.11\%         & 27.85\%         & 44.96\%   &  9.74\%      \\
            4    & CLIP$\dagger$~\cite{fu2022there}        & Fine-tune    & 15.72\%         & 37.13\%         & 49.74\%   &  13.82\%       \\
            5    & CLIP+Seg~\cite{fu2022there}    & Fine-tune    & 16.46\%         & 37.48\%         & 50.52\%    &  14.63\%  \\
            6 & \textbf{QR-CLIP} (Ours) & Fine-tune & { \textbf{19.51\%}} & { \textbf{38.48\%}} & { \textbf{51.25\%}}   & { \textbf{17.65\%}} \\ \hline
            \multicolumn{3}{l||}{\cellcolor[HTML]{DAE8FC} \textit{Relative Improvements (AVG: 10.82\%)}} & \cellcolor[HTML]{EFEFEF}\textit{+18.55\%} & \cellcolor[HTML]{EFEFEF}\textit{+2.67\%} & \cellcolor[HTML]{EFEFEF}\textit{+1.45\%} & \cellcolor[HTML]{EFEFEF}\textit{+20.62\%}\\ 
            \hline\hline
            \multicolumn{7}{c}{\cellcolor[HTML]{DAE8FC}Time Reasoning} \\ \hline
            7    & ResNet-50~\cite{he2016deep}   & Fine-tune  & 0.84\%          & 5.14\%          & 39.99\%   & 0.46\%        \\
            8    & Swin-T~\cite{liu2021swin}      & Fine-tune  & 0.97\%          & 5.53\%          & 43.95\%  & 0.72\%      \\
            9    & CLIP~\cite{radford2021learning}       & Zero-Shot   & 0.46\%          & 2.42\%          & 39.90\%  & 0.25\%       \\
            10    & CLIP$\dagger$~\cite{fu2022there}     & Fine-tune    & 1.00\%          & 2.99\%          & 43.09\%  & 0.54\%       \\
            11    & CLIP+Seg~\cite{fu2022there}    & Fine-tune    & 0.92\%          & 3.15\%          & 42.89\%     & 0.71\%    \\
            12 & \textbf{QR-CLIP} (Ours) & Fine-tune & { \textbf{3.45\%}}  & { \textbf{10.97\%}} & { \textbf{50.53\%}} & { \textbf{1.49\%}}\\ \hline
            \multicolumn{3}{l||}{\cellcolor[HTML]{DAE8FC} \textit{Relative Improvements (AVG:116.32\%)}} & \cellcolor[HTML]{EFEFEF}\textit{+245\%} & \cellcolor[HTML]{EFEFEF}\textit{+98.4\%} & \cellcolor[HTML]{EFEFEF}\textit{+14.98\%} & \cellcolor[HTML]{EFEFEF}\textit{+106.9\%}\\ 
            \hline
        \end{tabular}
    }
    \vspace{-11pt}
\end{table*}

\subsection{Comparative Results}

\textbf{Location Reasoning.}
We compare the results of \textbf{QR-CLIP} with other methods for location reasoning in Table~\ref{tab:ModelSummary}. In this experiment, both \textbf{ResNet-50} and \textbf{Swin-T} models were initialized with ImageNet~\cite{deng2009imagenet} pre-trained weights and subsequently fine-tuned for location and time reasoning tasks using the TARA dataset with an additional classification head.
Our \textbf{QR-CLIP} model achieves an accuracy of 19.51\% (Rank@1). Additionally, it attains an Example-F1 score of 51.25\% for the hierarchical labels.
All the results collectively show that our method outperforms other methods.

\textbf{(1)} Compared with ResNet-50~\cite{he2016deep} and Swin-T~\cite{liu2021swin}, vanilla CLIP achieves an absolute improvement of 7.93\% and 4.41\% in location reason accuracy \textbf{(IDs: 1, 2, 3)}. 
It is evident that, in comparison to the vision model only trained on ImageNet~\cite{deng2009imagenet}, CLIP already possesses a certain level of knowledge for reasoning.
Meanwhile, our \textbf{QR-CLIP} achieves a significant advantage with 16.33\% and 12.81\% absolute improvements in terms of accuracy \textbf{(IDs: 1, 2, 6)}. 
These results show that conventional image classification methods are insufficient in inferring the abstract information conveyed by the images. 
While the CLIP model trained on large-scale internet data have the ability to identify locations based on image data. \textbf{QR-CLIP} significantly enhances this capability.

\textbf{(2)} 
Besides, compared to CLIP$\dagger$ and the previous SOTA method CLIP+Seg, \textbf{QR-CLIP} exhibits a significant accuracy absolute improvement of 3.79\% and 3.05\%, respectively, along with a corresponding increase in F1-Score by 3.83\% and 3.02\%, respectively \textbf{(IDs: 4, 5, 6)}. 
Other evaluation metrics also improved. The results show that \textbf{QR-CLIP} can effectively utilize OWK to establish a stronger connection between image and location information through fine-tuning CLIP. However, we have observed that the improvement in Example-F1 is not as apparent. 
We argue that this is because of the mechanism of Example-F1. To illustrate,   consider the image shown in Fig.~\ref{fig:pipeline}, which contains many elements of Arabia, such as turbans and Arabic writing.
It is not difficult for many models to recognize that this image was captured in the Middle East and to predict its hierarchical label as \{`\emph{Asia}'\}. However, they failed when asked to predict the entire label \emph{Riyadh, Saudi Arabia, Asia}. Therefore, the discrepancy in other metrics may be more noticeable. 

\textbf{Time Reasoning.} Table~\ref{tab:ModelSummary} also presents the performance of our method and existing techniques for time reasoning. 
The Accuracy (Rank@1) of \textbf{QR-CLIP} is 3.45\%, and Example-F1 is 50.53\%; compared to the CLIP model, the two metrics have been absolutely improved by 2.99\% and 10.63\%, respectively \textbf{(IDs: 9, 12)}. 
Compared with CLIP$\dagger$ and CLIP+Seg, which are also based on fine-tuned CLIP, our method achieves absolute improvements of 2.45\% and 2.53\%  in time reasoning accuracy, respectively. 
Compared with traditional image classification methods, \textbf{QR-CLIP} exhibits absolute advantages in all metrics \textbf{(IDs: 7, 8, 12)}.
In addition, the image lacks time-related information, resulting in a fine-tuned CLIP method accuracy of only about 1\% for image time reasoning, which is significantly lower than the accuracy of location reasoning \textbf{(IDs: 10, 11)}. 

It is not surprising that even for humans, determining the time a photo was taken can be difficult, as illustrated by the sample image in Fig.~\ref{fig:pipeline}. For instance, if one is unfamiliar with Cristiano Ronald or lacks specific knowledge, they may not recognize that the time stamp on the image, \{`\emph{03-01-2023}'\}
indicates the date the photo was taken. Nevertheless, the method proposed in this paper is highly effective, as our model achieves a relative lift of +245.00\% for predicting time and significantly narrows the gap with location reasoning.

\subsection{Ablation Study}
\textbf{Analysis on Additional \texttt{[CLS]}.} Following the network design process, all experiments in this section were conducted with only step \textcolor{red}{1} in the Quantity Module (Sec~\ref{sec:qm}). As shown in Table~\ref{tab:Additional CLS}, both of different \texttt{[CLS]} aggregation methods and varying numbers of \texttt{[CLS]} can impact network performance. 

Firstly, we observe differences in network performance related to \texttt{[CLS]} aggregation methods. 
The $[\texttt{CLS}^*]_i^v$ approach combines all additional \texttt{[CLS]} tokens using MLPs to compute similarity with location or time labels. In contrast, $[\texttt{CLS}]_i^v$ computes the similarity between each \texttt{[CLS]} token and the labels, and the final matching pairs are obtained by taking the average of these similarities.

When comparing \texttt{[CLS$_i^*$]} and $[\texttt{CLS}]_i^v$ with the same number (\emph{i.e.,} $n=2$) of \texttt{[CLS]}, the latter exhibits 7.92\% and 0.77\% higher accuracy in location and time reason \textbf{(IDs: 13, 16, 20, 23)}. Moreover, the performance of \texttt{[CLS$_i^*$]} is not significantly impacted by the number of \texttt{[CLS]} \textbf{(IDs: 13-15, 20-22)}. 
We argue that utilizing \texttt{MLP} for \texttt{[CLS]} aggregation has the potential to disrupt the original representation of CLIP. The \texttt{[CLS]} embeddings of CLIP, which are obtained through extensive pretraining, already show a comprehensive understanding of both images and text. The incorporation of further non-linear transformations through \texttt{MLP} aggregation may lead to the loss of certain original information.
Instead of directly comparing $[\texttt{CLS}]_i^v$ with location and time labels, a more effective approach is to independently calculate the similarities between each $[\texttt{CLS}]_i^v$ and the labels. The final reasoning can then be obtained by averaging the calculated similarities.
This strategy enables better utilization of the advantages offered by pre-trained vision-language models while preserving their original representations.

Then we analyze how different numbers of \texttt{[CLS]} affect the model's performance.
No significant differences in performance were observed when $n=4$ and $6$. However, when $n$ was increased to 8, the model's performance exhibited a decrease.
So we utimately chose $n=6$ in the following experiments \textbf{(IDs: 17-19, 24-26)}. 
This decision strikes a balance between leveraging the advantages of additional \texttt{[CLS]} embeddings and avoiding potential redundancy, effectively optimizing the overall model performance. 
The results indicate that the additional \texttt{[CLS]}  effectively increases image cues by constructing multiple perspectives, which has promising benefits.

\begin{table}[t!]
    \centering
    \caption{Performance of additional \texttt{[CLS]} in \textbf{QR-CLIP} with different number and reasoning methods. $[\texttt{CLS}^*]_i^v$ aggregates all additional \texttt{[CLS]} tokens through MLPs to compute the similarity. On the other hand, $[\texttt{CLS}]_i^v$ calculates the label similarity for each pair individually. Here, $n$ represents the number of \texttt{[CLS]} tokens.
    }
    \label{tab:Additional CLS}
    \resizebox{\linewidth}{!}{
        \setlength\tabcolsep{3pt}
        \begin{tabular}{ll||c|c|c}
            \hline
            \cellcolor[HTML]{CBCEFB}ID   & \cellcolor[HTML]{CBCEFB}Method  & \cellcolor[HTML]{CBCEFB}Accuracy (Rank@1)     & \cellcolor[HTML]{CBCEFB}Rank@5  & \cellcolor[HTML]{CBCEFB}Example-F1 \\ \hline\hline
            \multicolumn{5}{c}{\cellcolor[HTML]{DAE8FC}Location Reasoning} \\ \hline
            13 & CLIP+$[\texttt{CLS}^*]_i^v$($n$=2) & 8.99\%    & 27.13\%   & 43.21\%     
            \\
            14 & CLIP+$[\texttt{CLS}^*]_i^v$($n$=4)  & 9.11\%   & 27.03\%  & 43.74\% 
            \\
            15 & CLIP+$[\texttt{CLS}^*]_i^v$($n$=6)  & 8.52\%    & 26.88\%  & 42.95\%      
            \\
            \hline\hline
            16 & CLIP+$[\texttt{CLS}]_i^v$($n$=2)    & 16.91\%     & 37.91\%    & 49.47\%         \\
            17 & CLIP+$[\texttt{CLS}]_i^v$($n$=4)    & \textbf{17.53\%}     & \textbf{38.10\%}    & 50.03\%         \\
            18 & CLIP+$[\texttt{CLS}]_i^v$($n$=6)    & 17.47\%     & 38.06\%    & \textbf{50.10\%}         \\
            19 & CLIP+$[\texttt{CLS}]_i^v$($n$=8)    & 16.78\%     & 37.40\%    & 48.71\%         \\
            \hline\hline
            \multicolumn{5}{c}{\cellcolor[HTML]{DAE8FC}Time Reasoning} \\ \hline
            20 & CLIP+$[\texttt{CLS}^*]_i^v$($n$=2)  & 1.13\%  &  3.01\%    & 43.72\%        
            \\
            21 & CLIP+$[\texttt{CLS}^*]_i^v$($n$=4)  & 1.01\%  &  2.93\%     & 43.37\%  
            \\
            22 & CLIP+$[\texttt{CLS}^*]_i^v$($n$=6)  & 1.22\%  &  2.97\%     & 43.85\%        
            \\
            \hline\hline
            23 & CLIP+$[\texttt{CLS}]_i^v$($n$=2)    & 1.90\%    & 5.25\%   & 45.62\%         
            \\
            24 & CLIP+$[\texttt{CLS}]_i^v$($n$=4)    & 1.99\%    & \textbf{5.38\%}   & 45.68\% 
            \\
            25 & CLIP+$[\texttt{CLS}]_i^v$($n$=6)    & \textbf{2.03\%}    & 5.33\%   & \textbf{45.72\%}  
            \\
            26 & CLIP+$[\texttt{CLS}]_i^v$($n$=8)    & 1.66\%    & 5.16\%   & 45.27\%      
            \\
            \hline
        \end{tabular}
    }
    \vspace{-11pt}
\end{table}

\begin{table}[t!]
\caption{The impact of various loss functions and components on performance. $LL$, $GL$ indicate the local loss and global loss, respectively. \textbf{QR-CLIP} means the model contains entirely Quantity Module (QM: Sec.~\ref{sec:qm}) and Relevance Module (RM: Sec.~\ref{sec:rm}).}
\label{tab:loss function} 
\centering
\resizebox{\linewidth}{!}{%
\setlength\tabcolsep{1pt}
\begin{tabular}{ll||c|c|c}
\hline
\cellcolor[HTML]{CBCEFB}ID   & \cellcolor[HTML]{CBCEFB}Method  & \cellcolor[HTML]{CBCEFB}Accuracy (Rank@1)     &\cellcolor[HTML]{CBCEFB} Rank@5  & \cellcolor[HTML]{CBCEFB}Example-F1    \\ \hline\hline
\multicolumn{5}{c}{\cellcolor[HTML]{DAE8FC}Location Reasoning (Only QM)}                                           \\ \hline
27 & CLIP+$[\texttt{CLS}]_i^v$($n$=6)+$LL$  & 16.56\% & 37.08\%  &     49.85\%
\\
28 & CLIP+$[\texttt{CLS}]_i^v$($n$=6)+$GL$  &  17.04\% & 37.26\% &  49.93\% 
\\
29 & CLIP+$[\texttt{CLS}]_i^v$($n$=6)+$LL$+$GL$  & \textbf{17.47\%} & \textbf{38.00\%}  & \textbf{50.10\%}  
\\
 \hline\hline
 \multicolumn{5}{c}{\cellcolor[HTML]{DAE8FC}Time Reasoning (Only QM)}                                       \\ \hline
30 & CLIP+$[\texttt{CLS}]_i^v$($n$=6)+$LL$  & 1.31\%  & 5.56\%  &    44.83\% 
\\
31 & CLIP+$[\texttt{CLS}]_i^v$($n$=6)+$GL$  & 1.84\%  & 5.77\%   &  44.53\% 
\\
32 & CLIP+$[\texttt{CLS}]_i^v$($n$=6)+$LL$+$GL$  & \textbf{2.03\%}   & \textbf{6.33\%}  & \textbf{45.72\%}   
\\
 \hline
\multicolumn{5}{c}{\cellcolor[HTML]{DAE8FC}Location Reasoning (QR-CLIP: QM+RM)}                                           \\ \hline
33 & CLIP+$[\texttt{CLS}]_i^v$($n$=6)+$LL$  & 19.19\% & 37.12\%  &  50.63\%
\\
34 & CLIP+$[\texttt{CLS}]_i^v$($n$=6)+$GL$  & 18.61\% & 37.49\% &  50.91\% 
\\
35 & CLIP+$[\texttt{CLS}]_i^v$($n$=6)+$LL$+$GL$  & \textbf{19.51\%} & \textbf{38.48\%}  & \textbf{51.25\%}  
\\
 \hline\hline
 \multicolumn{5}{c}{\cellcolor[HTML]{DAE8FC}Time Reasoning (QR-CLIP: QM+RM)}                                       \\ \hline
36 & CLIP+$[\texttt{CLS}]_i^v$($n$=6)+$LL$  & 3.12\%  & \textbf{11.49\%}  &    48.85\% 
\\
37 & CLIP+$[\texttt{CLS}]_i^v$($n$=6)+$GL$  & 2.78\%  & 9.86\%   &  47.11\% 
\\
38 & CLIP+$[\texttt{CLS}]_i^v$($n$=6)+$LL$+$GL$  & \textbf{3.45\%}   & 10.97\%  & \textbf{50.53\%}  
\\
 \hline
\end{tabular}
}
\vspace{-11pt}
\end{table}

\begin{table}[t!]
    \centering
    \caption{The results of the effect of increasing the candidate numbers of OWK.}
    \label{tab:impact of Open-World Knowledge}
    \resizebox{\linewidth}{!}{
        \setlength\tabcolsep{6pt}
        \begin{tabular}{ll||c|c|c}
            \hline
            \cellcolor[HTML]{CBCEFB}ID & \cellcolor[HTML]{CBCEFB}Candidate OWK & \cellcolor[HTML]{CBCEFB}Accuracy (Rank@1) & \cellcolor[HTML]{CBCEFB}Rank@5 & \cellcolor[HTML]{CBCEFB}Example-F1 \\ \hline\hline
            \multicolumn{5}{c}{\cellcolor[HTML]{DAE8FC}Location Reasoning} \\ \hline
            39 & 29,243 & 18.26\% & 37.97\% & 50.29\% \\
            40 & 52,159 & 18.83\% & 38.29\% & 50.45\% \\
            41 & 122,408 & \textbf{19.51\%} & \textbf{38.48\%} & \textbf{51.25\%} \\
            \hline\hline
            \multicolumn{5}{c}{\cellcolor[HTML]{DAE8FC}Time Reasoning} \\ \hline
            42 & 29,243 & 2.26\% & 6.77\% & 47.85\% \\
            43 & 52,159 & 2.88\% & 10.67\% & 48.52\% \\
            44 & 122,408 & \textbf{3.45\%} & \textbf{10.97\%} & \textbf{50.53\%} \\
            \hline
        \end{tabular}
    }
    \vspace{-11pt}
\end{table}

\begin{table}[t!]
    \centering
    \caption{The effect of different scoring mechanisms on network performance, where Score$_v$ indicates that only images are scored and Score$_t$ means scoring OWK only.}
    \label{tab:Scoring Mechanisms}
    \resizebox{\linewidth}{!}{
        \setlength\tabcolsep{6pt}
        \begin{tabular}{ll||c|c|c}
            \hline
            \cellcolor[HTML]{CBCEFB}ID   & \cellcolor[HTML]{CBCEFB}Method  & \cellcolor[HTML]{CBCEFB}Accuracy (Rank@1)     &\cellcolor[HTML]{CBCEFB} Rank@5  & \cellcolor[HTML]{CBCEFB}Example-F1 \\
            \hline\hline
            \multicolumn{5}{c}{\cellcolor[HTML]{DAE8FC}Location Reasoning} \\ \hline
            45 & Score$_v$  & 15.45\% & 35.63\%  & 47.87\%     
            \\
            46 & Score$_t$ & 18.56\%  & 37.55\% & 50.94\% 
            \\
            47 & Proposed  & \textbf{19.51\%} & \textbf{38.48\%}  & \textbf{51.25\%}  
            \\
            \hline\hline
            \multicolumn{5}{c}{\cellcolor[HTML]{DAE8FC}Time Reasoning} \\ \hline
            48 & Score$_v$  & 2.72\%  & 10.61\%  & 50.38\%       
            \\
            49 & Score$_t$  & 2.83\%  & 10.43\%  & 50.42\%
            \\
            50 & Proposed   & \textbf{3.45\%}   & \textbf{10.97\%}  & \textbf{50.53\%}  
            \\
            \hline
        \end{tabular}
    }
    \vspace{-11pt}
\end{table}

\begin{table}[t!]
    \centering
    \caption{The impact of Diversity in OWK on Relevance Module Performance. where 'Uniform Search' represents the approach of using the highest-scoring wiki entry for all searches, and 'Distinct Search' involves searching through diverse wiki entries with a zero duplication rate.}
    \label{tab:Diversity Analysis}
    \fontsize{25pt}{30pt}\selectfont
    \resizebox{\linewidth}{!}{
        \setlength\tabcolsep{6pt}
        \begin{tabular}{ll||c|c|c|c}
            \hline
            \cellcolor[HTML]{CBCEFB}ID   & \cellcolor[HTML]{CBCEFB}Method  &  \cellcolor[HTML]{CBCEFB}Duplication Rate   & \cellcolor[HTML]{CBCEFB}Accuracy (Rank@1)     &\cellcolor[HTML]{CBCEFB} Rank@5  & \cellcolor[HTML]{CBCEFB}Example-F1 \\
            \hline\hline
            \multicolumn{6}{c}{\cellcolor[HTML]{DAE8FC}Location Reasoning} \\ \hline
            51 & Uniform Search  & 100\% & 19.17\% & 38.11\%  & 50.97\%     
            \\
            52 & Distinct Search & 0\% & 18.88\%  & 37.97\% & 50.43\% 
            \\
            53 & Proposed  & 67.62\%  & \textbf{19.51\%} & \textbf{38.48\%}  &\textbf{51.25\%}  
            \\
            \hline\hline
            \multicolumn{6}{c}{\cellcolor[HTML]{DAE8FC}Time Reasoning} \\ \hline
            54 & Uniform Search & 100\% & 3.16\%  & 10.63\%  & 50.11\%       
            \\
            55 & Distinct Search & 0\% & 3.03\%  & 10.74\%  & 49.60\%
            \\
            56 & Proposed  & 67.10\% & \textbf{3.45\%}   & \textbf{10.97\%}  & \textbf{50.53\%} 
            \\
            \hline
        \end{tabular}
    }
    \vspace{-11pt}
\end{table}

\begin{figure*}[t!]
\begin{center}
\includegraphics[width=1.0\linewidth]{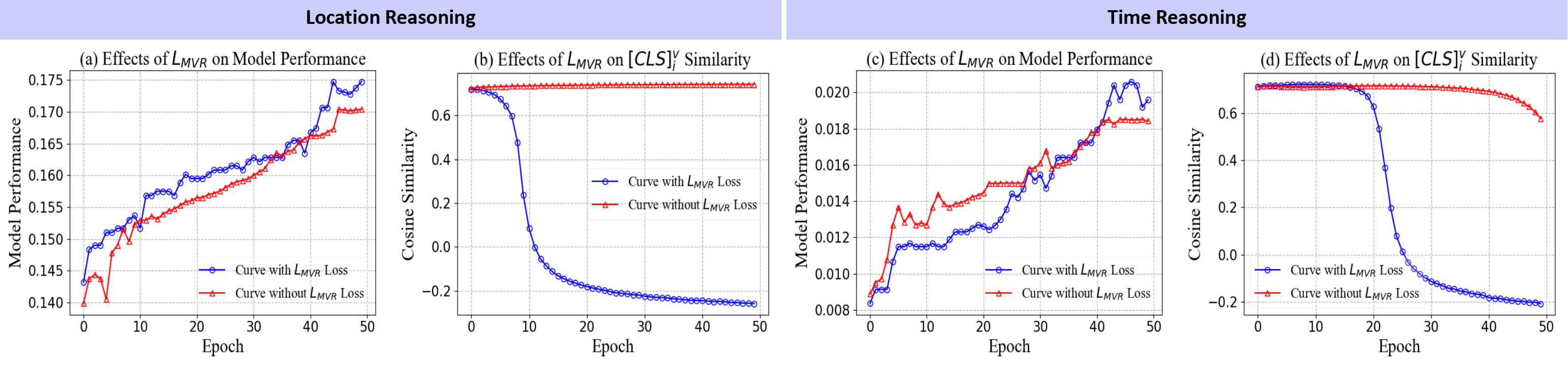}
\end{center}
\vspace{-15pt}
\caption{
    This figure illustrates the impact of diversity on the performance of the Quantity Module. Among them, graph (a) compares the model's performance in location reasoning tasks with the blue curve representing the model's performance with $L_{MVR}$ loss and the red curve representing the model's performance without $L_{MVR}$ loss. Graph (b) tracks the changes in cosine similarity among $[\texttt{CLS}]_i^v$ during training. Similarly, graphs (c) and (d) present the results of experiments in time reasoning tasks.}
\label{fig: Multiview Diversity Analysis}
\end{figure*}

\textbf{Effectiveness of Losses and Modules.}
We further analyze the impact of different loss functions, namely local loss ($LL$) and global loss ($GL$), as well as the contributions of the Quantity (Sec~\ref{sec:qm}) and Relevance Modules (Sec~\ref{sec:rm}) to the performance of the model.

The Table~\ref{tab:loss function} shows that the differential impact of loss functions on Quantity and Relevance Modules. The global loss enhances the performance of the Quantity Module, as evidenced by its superior accuracy over the local loss in location and time reasoning tasks \textbf{(IDs: 27 vs 28, 30 vs 31)}. 
For the Relevance Module, local loss performs better \textbf{(IDs: 33, 34, 36, 37)}. The combined use of both loss further improves model's effectiveness and complementarity \textbf{(IDs: 27-38)}, demonstrating their combined potential.

When comparing the Quantity Module to the entire \textbf{QR-CLIP}, we observe that the inclusion of the Relevance Module significantly improves the reasoning abilities \textbf{(IDs: 29, 32, 35, 38)},  which verifies that the overall designs of the two modules are reasonable.

\textbf{Impact of OWK.} 
To validate the impact of varying amounts of OWK in the environment, we conduct an experiment to determine whether increasing the number of OWK is beneficial.
As shown in Table~\ref{tab:impact of Open-World Knowledge}, the addition of 122,408 OWK resulted in more accurate reasons by the network (absolute lift by 2.04\% and 1.42\%) for location and time \textbf{(IDs: 18, 25, 41, 44)}, compared to the method without OWK. 
These results show that our method effectively utilizes OWK to enhance the accuracy of the model for image location and time. Besides, the performance gradually improves as the number of OWK increases  \textbf{(IDs: 39-44)}.  
It also shows that our method has the capability to explore a wider range of OWK. However, comparing each $[\texttt{CLS}]_i^v$ with $122,408$ OWK is already time-consuming and  limits the ability to increase the amount. In the future, we will strive to find a more efficient approach to overcome this challenge.

\textbf{Performance of Scoring Mechanism.} 
This section evaluates the performance of different scoring mechanisms in the Relevance Module (Sec~\ref{sec:rm}), and the experimental results are shown in Table~\ref{tab:Scoring Mechanisms}.
When utilizing Score$_v$, certain image features may be weakened, and the accuracy of time and location reasoning may decrease after fusing OWK \textbf{(IDs: 45, 47, 48, 50)}.
When using the scoring mechanism on text (Score$_t$), only OWK was considered during the fusion process. As a result, the accuracy of location and time reasoning absolutely improved by 3.11\% and 0.11\%, respectively \textbf{(IDs: 45, 46, 48, 49)}. 
This suggests that the weights exert a significant influence on the final reasoning.  
When both image and OWK embeddings are scored, the accuracy of location and time reasoning increases by 4.06\% and 0.73\%, respectively \textbf{(IDs: 45, 47, 48, 50)}. 
It is evident that providing essential information during interactions supports \textbf{QR-CLIP}'s ability to comprehend the abstract concepts behind images, which aligns with the distributed cognition theory.

\textbf{Multiview Diversity Analysis.}  
To investigate the impact of the diversity of $[\texttt{CLS}]_i^v$ in the Quantity Module on model performance, we conducted an experiment using $L_{MVR}$ loss (Eq.~\ref{MVR_loss}) as the variable.

As shown in Fig.~\ref{fig: Multiview Diversity Analysis}, the incorporation of $L_{MVR}$ leads to a peak accuracy of 17.47\% (graph(a): \textcolor{blue}{blue curve}), surpassing the model without $L_{MVR}$, which achieves an accuracy of 17.04\% (graph(a): \textcolor{red}{red curve}). In addition, the inclusion of $L_{MVR}$ results in a decrease in the similarity between each $[\texttt{CLS}]_i^v$ (graph(b): \textcolor{blue}{blue curve}). Similar trends are also observed in the experimental results of time reasoning tasks (graph(c), graph(d)).
These results show that the introduction of $L_{MVR}$ enhances the dissimilarity between each $[\texttt{CLS}]_i^v$, allowing the model to effectively differentiate and capture a wide range of perspectives, thereby improving model performance.

For the Relevance Module, we evaluated the effect of diversity on performance using search-derived OWK. As seen in Table~\ref{tab:Diversity Analysis}, Uniform Search achieved only 19.17\% in Location Reasoning \textbf{(ID: 51)}, despite using top-scoring OWK, due to lack of diversity. Distinct Search, with six unique OWK, reached 18.88\% \textbf{(ID: 52)}. A similar pattern was observed in Time Reasoning \textbf{(IDs: 54-56)}.

The proposed method assigns unique scores to each $[\texttt{CLS}]_i^v$ and their corresponding OWK, leading to superior performance in both tasks. It achieves a desirable equilibrium between diversity and reasoning capability, even in the presence of about 67\% duplication in OWK. Conversely, Uniform and Distinct Searches, which prioritize consistency and diversity, respectively, resulted in lower performance.

\subsection{Method Generalization Validation}
\textbf{Testing on Different Datasets.} In this section, we evaluated the generalization performance of \textbf{QR-CLIP} on three different datasets. TARA-Dev~\cite{fu2022there} contains 1,552 images distinct from the test set. 
TARA-Interest~\cite{fu2022there} comprises 30 images related to news events occurring after January 2021, which is the cut-off date for the CLIP model. These images are specifically designed to test the model's generalization ability rather than its memorization ability. The COFAR dataset~\cite{cofar2022} includes landmark images and their corresponding descriptions. It is noteworthy that all experiments are conducted without any additional training on the corresponding datasets.

\begin{figure}[t!]
\begin{center}
\includegraphics[width=1.0\linewidth]{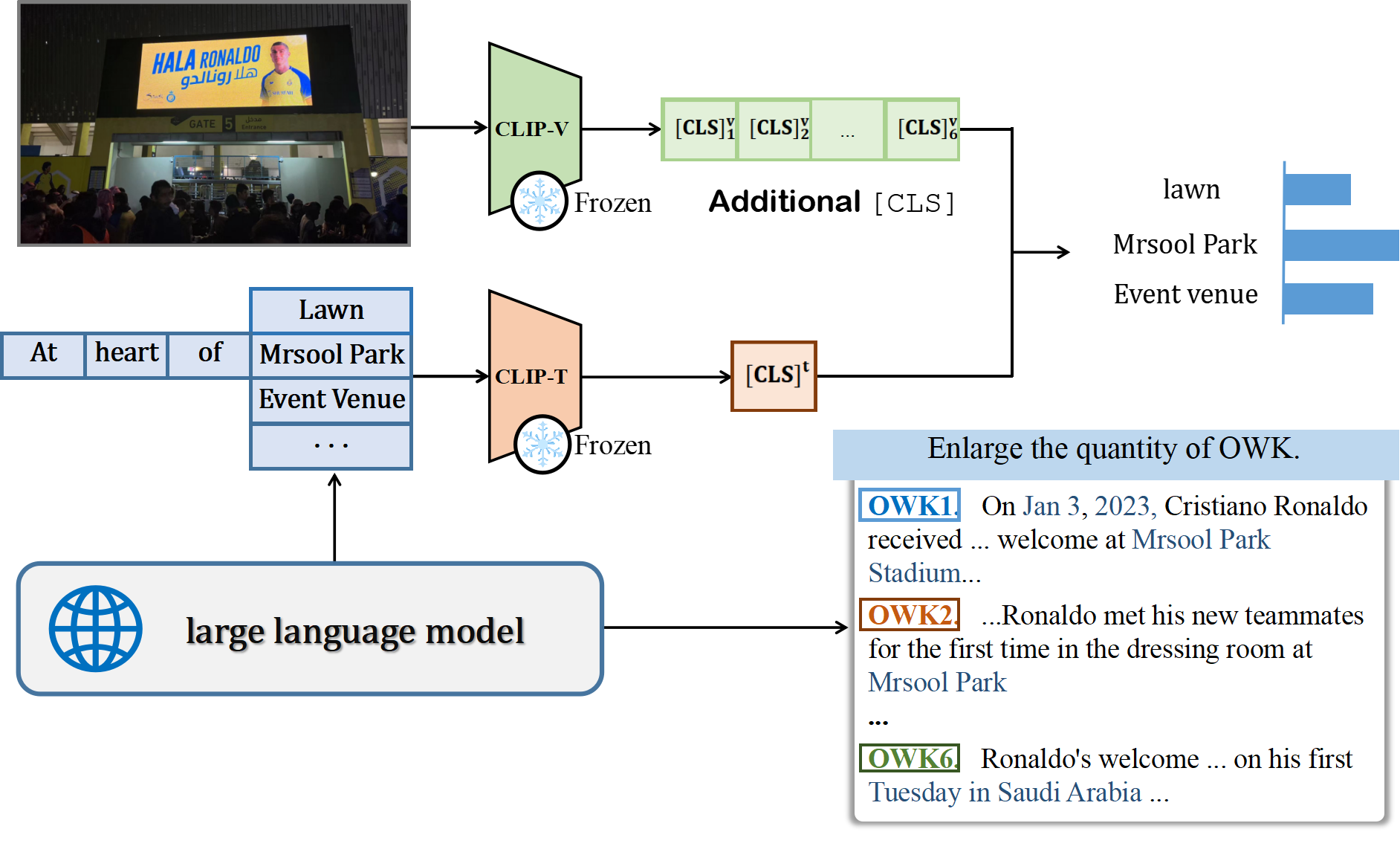}
\end{center}
\vspace{-15pt}
    \caption{
    The illustration of searching OWK from large language models. The CLIP-V outputs diverse representations of image location and time as $[\texttt{CLS}]_i^v$, which enables semantic alignment between generated results and input images. This alignment facilitates the selection of the most suitable token for OWK search based on visual information. Each $[\texttt{CLS}]_i^v$ yields a corresponding textual knowledge, consistent with step \textcolor{red}{2} in Fig.~\ref{fig:model}.
    }
\label{fig:clip+gpt2}
\end{figure}

\begin{table}[t!]
\caption{Performance comparison of location and time reasoning tasks across different datasets. Here, we record the results of TARA-Dev~\cite{fu2022there}, TARA-Interest~\cite{fu2022there} and COFAR~\cite{cofar2022}.}
\label{tab: Generalization Validation} 
\centering
\resizebox{\linewidth}{!}{%
\setlength\tabcolsep{3.4pt}
\begin{tabular}{ll||c|c|c}
\hline
\cellcolor[HTML]{CBCEFB}ID   & \cellcolor[HTML]{CBCEFB}Method  & \cellcolor[HTML]{CBCEFB}Accuracy (Rank@1)     &\cellcolor[HTML]{CBCEFB} Rank@5  & \cellcolor[HTML]{CBCEFB}Example-F1    \\ \hline\hline
\multicolumn{5}{c}{\cellcolor[HTML]{D5F0FF}TARA-Dev~\cite{fu2022there}} \\ 
\hline
\multicolumn{5}{c}{\cellcolor[HTML]{DAE8FC}Location Reasoning}                                           \\ \hline

57    & CLIP~\cite{radford2021learning}        & 10.99\%         & 29.72\%         & 45.90\%       \\
58    & CLIP+Seg~\cite{fu2022there}            & 15.88\%         & 39.15\%         & 51.83\%        \\
59    & \textbf{QR-CLIP} (Ours)        & { \textbf{20.35\%}}  & { \textbf{40.23\%}}     & { \textbf{51.60\%}}   \\
 \hline\hline
 \multicolumn{5}{c}{\cellcolor[HTML]{DAE8FC}Time Reasoning}                                       \\ \hline
60    & CLIP~\cite{radford2021learning}        & 0.53\%         & 1.82\%         & 42.14\%       \\
61    & CLIP+Seg~\cite{fu2022there}            & 0.53\%         & 2.50\%         & 43.55\%        \\
62    & \textbf{QR-CLIP} (Ours)        & { \textbf{6.53\%}}  & { \textbf{18.89\%}}     & { \textbf{53.26\%}}   \\
 \hline \hline
\multicolumn{5}{c}{\cellcolor[HTML]{D5F0FF}TARA-Interest~\cite{fu2022there} } \\
 \hline 
\multicolumn{5}{c}{\cellcolor[HTML]{DAE8FC}Location Reasoning}                                           \\ \hline
63    & CLIP~\cite{radford2021learning}        & 13.33\%         & 27.85\%         & 56.44\%       \\
64    & CLIP+Seg~\cite{fu2022there}            & 23.33\%         & 37.48\%         & 63.11\%        \\
65    & \textbf{QR-CLIP} (Ours)        & { \textbf{58.62\%}}  & { \textbf{86.20\%}}     & { \textbf{80.46\%}}   \\
 \hline\hline
 \multicolumn{5}{c}{\cellcolor[HTML]{DAE8FC}Time Reasoning}                                       \\ \hline
66    & CLIP~\cite{radford2021learning}        & 0.00\%         & 1.85\%         & 24.56\%       \\
67    & CLIP+Seg~\cite{fu2022there}            & 3.33\%         & 9.48\%         & 24.43\%        \\
68    & \textbf{QR-CLIP} (Ours)        & { \textbf{20.69\%}}  & { \textbf{41.38\%}}     & { \textbf{60.34\%}}   \\
 \hline
\multicolumn{5}{c}{\cellcolor[HTML]{D5F0FF} COFAR~\cite{cofar2022} } \\
\hline
\multicolumn{5}{c}{\cellcolor[HTML]{DAE8FC}Location Reasoning}                                           \\ \hline

69    & CLIP~\cite{radford2021learning}        & 70.96\%         & 84.29\%         & 81.97\%       \\
70    & CLIP+Seg~\cite{fu2022there}            & 70.00\%         & 83.33\%         & 80.05\%        \\
71    & \textbf{QR-CLIP} (Ours)        & \textbf{71.42\%} & { \textbf{85.71\%}}     & { \textbf{85.14\%}}   \\
 \hline
\end{tabular}}
\vspace{-11pt}
\end{table}

\begin{table}[t!]
    \centering
    \caption{The impact results of different OWK sources on location and time reasoning tasks. The OWK sources utilized include None (\textit{N/A}), Wikipedia, and a large language model (\textit{GPT-2}).}
    \label{tab:search form LLM}
    \fontsize{25pt}{30pt}\selectfont
    \resizebox{\linewidth}{!}{
        \setlength\tabcolsep{6pt}
        \begin{tabular}{lc||c|c|c|c}
            \hline
            \cellcolor[HTML]{CBCEFB}ID   &
            \cellcolor[HTML]{CBCEFB}OWK Source   &
            \cellcolor[HTML]{CBCEFB}Method  & \cellcolor[HTML]{CBCEFB}Accuracy (Rank@1)     &\cellcolor[HTML]{CBCEFB} Rank@5  & \cellcolor[HTML]{CBCEFB}Example-F1 \\
            \hline\hline
            \multicolumn{6}{c}{\cellcolor[HTML]{DAE8FC}Location Reasoning} \\ \hline
            72 & \textit{N/A} & CLIP  & 15.72\% & 37.13\%  & 49.74\%     
            \\
            73 & \textit{N/A} & CLIP+Seg & 16.46\%  & 37.48\% & 50.52\% 
            \\
            74 & \textit{Wikipedia} &QR-CLIP  & \textbf{19.44\%} & \textbf{38.48\%}  & \textbf{51.25\%} 
            \\
            75 & \textit{GPT-2} &QR-CLIP  & 19.17\% & 38.13\%  & 50.98\% 
            \\
            \hline\hline
            \multicolumn{6}{c}{\cellcolor[HTML]{DAE8FC}Time Reasoning} \\ \hline
            76 & \textit{N/A} & CLIP  & 1.0\% & 2.99\%  & 43.09\%     
            \\
            77 & \textit{N/A} & CLIP+Seg & 0.92\%  & 3.15\% & 42.89\% 
            \\
            78 & \textit{Wikipedia} &QR-CLIP  & \textbf{3.45\%} & \textbf{10.97\%}  & \textbf{50.53\%} 
            \\
            79 & \textit{GPT-2} &QR-CLIP  & 3.31\% & 8.78\%  & 45.96\% 
            \\
            \hline
        \end{tabular}
    }
    \vspace{-11pt}
\end{table}

As Illustrated in Table~\ref{tab: Generalization Validation}, the experimental results show the efficacy of our \textbf{QR-CLIP} model in comparison to other methods. 
On the TARA-Dev dataset, \textbf{QR-CLIP} achieved an accuracy of 20.35\% in location reasoning and 6.53\% in time reasoning, surpassing both the CLIP and CLIP+Seg models \textbf{(IDs: 57-59, 60-62)}. 
These results validate the effectiveness of the model in handling diverse and previously unseen images. A similar trend was observed in the TARA-Interest dataset, where \textbf{QR-CLIP} attained an accuracy of 58.62\% \textbf{(ID: 65)} in location reasoning and 20.69\% \textbf{(ID: 68)} in time reasoning. 
This not only shows that the generalization capability of \textbf{QR-CLIP}, but also suggests its potential uses in evolving real-world scenarios. Furthermore, on the COFAR dataset, \textbf{QR-CLIP} once again outperforms other models with a location reasoning accuracy of 71.42\% \textbf{(ID: 71)}. 
Given that landmark descriptions contain more context information, our model has achieved significant growth in all metrics concerning location reasoning. This result  corroborates its robustness and adaptability of our model to different data types and tasks.

In summary, our method successfully motivates visual language models to perform higher-level reasoning while maintaining the potent generalization capabilities of large-scale pre-training models. As a result, this research not only provides substantial support for the further studies and application of location and time reasoning tasks, but also carries positive implications for enhancing the generalization performance of large-scale visual language models in specific tasks.

\textbf{Search Knowledge From Large Language Model.} Existing large language models have accumulated a vast amount of implicit knowledge through  training on extensive text data.
The knowledge encoded in the model can be accessed through specific querying and decoding operations. We propose utilizing these large-scale language models as an OWK base to further verify the generalization capacity of our method.

\begin{figure*}[t!]
\begin{center}
\vspace{-7pt}
\includegraphics[width=.98\linewidth]{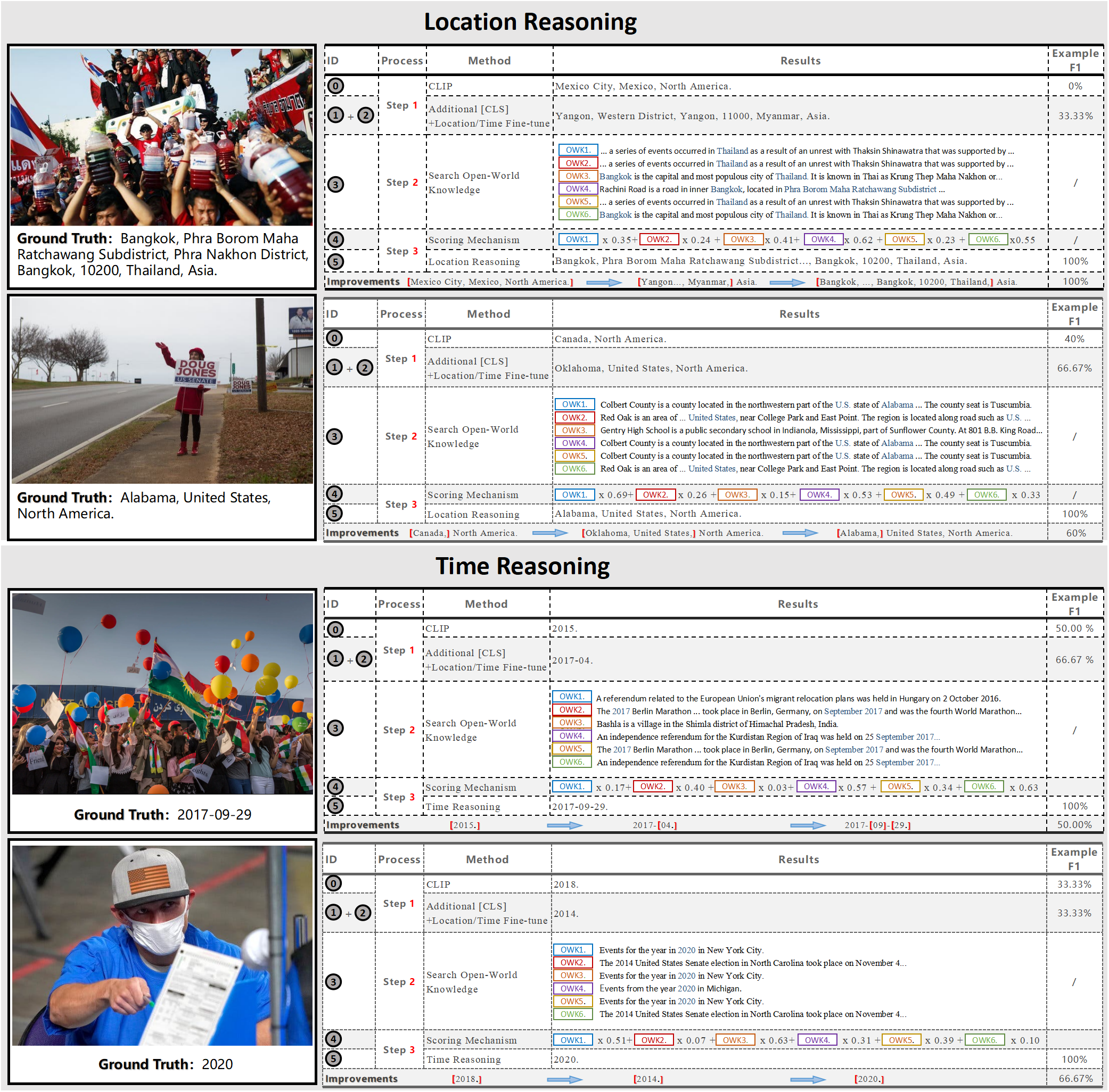}
\end{center}
\vspace{-10pt}
 \caption{
 We show visualizations of 5 procedures involved in \textbf{QR-CLIP}. For each process, readers can refer to Fig.~\ref{fig:model}.
 }\label{fig:vis}
\end{figure*}

This experiment utilizes the GPT-2 language model~\cite{radford2019language}, as shown in the Fig.~\ref{fig:clip+gpt2}. To familiarize the model with the text distribution relevant to the task, we employed 29,243 candidate OWK \textbf{(ID: 39)} for the post-training of GPT-2, which only required two hours of runtime on a single NVIDIA GeForce RTX 3090. The experiment employs the Magic~\cite{su2022language} and Simtcg~\cite{su2022contrastive} decoding methods. By integrating the similarity between the token and the image generated at each step by the language model into the decoding score, we execute a zero-shot knowledge search from the large language model.

The experimental findings illustrate that the incorporation of GPT-2 as an OWK source enhances the model's capabilities in location and time reasoning tasks \textbf{(IDs: 75, 79)}, even though its performance in these domains trails behind the model that leverages Wikipedia as an OWK \textbf{(IDs: 74 vs 75, 78 vs 79)}. This confirms the potential of language models as sources of OWK. Moreover, it also suggests that explicit knowledge bases, like Wikipedia, which are meticulously curated and organized by humans, could be more beneficial in enhancing the reasoning capabilities of models due to their ability to provide more context-specific details.

The application of OWK, in both its explicit (Wikipedia) and implicit (GPT-2) forms, shows an improvement in the performance of both location and time reasoning tasks compared to scenarios without any OWK \textbf{(IDs: 72-75, 76-79)}. 
This highlights the significance of open knowledge and showcases the versatility of our approach. We effectively leverage various types of OWK in the environment to enhance the model's cognitive and reasoning abilities.

\subsection{Visualization}
We present several visual demonstrations for \textbf{QR-CLIP} in Figure~\ref{fig:vis}. The first figure shows the model's performance in a location reasoning task. Our \textbf{QR-CLIP} demonstrates significant improvements compared to vanilla CLIP~\cite{radford2021learning}, which served as a baseline and achieved lower Example-F1 scores (0\%). \textbf{QR-CLIP} utilizes image search to obtain OWKs containing location information related to visual content, such as Thailand. The scoring mechanism assigns weights to each OWK, favoring those rich in valuable location details, thus guiding the model to emphasize the most relevant location information.

In the third picture, we explore the application of \textbf{QR-CLIP} to time reasoning. Here, vanilla CLIP serves as the baseline and achieves lower Example-F1 scores (50.00\%).
However, after using additional $[\texttt{CLS}]$ and fine-tuning them using global and local losses, our \textbf{QR-CLIP} detects an image from different perspectives and get higher scores (66.67\%). Subsequently, \textbf{QR-CLIP} retrieves six OWK used as language input, all of which describe the abstract information expressed in the image content: a public participation activity.
In addition, each piece of knowledge contains a wealth of information regarding the time associated with the activity. The scoring mechanism assigns varying weights to each OWK, with the OWK lacking valuable time information receiving a lower weight. This guides the model to focus on the correct time information.

\section{Conclusion and Future Work}

In our study, we developed \textbf{QR-CLIP}, a model inspired by Hutchins's distributed cognition theory, for image-based location and time reasoning tasks. It contains two Modules. The \textbf{Quantity} Module enhances cognitive abilities by providing a suite of cognitive tools aimed at aggregating a maximal amount of open-world knowledge from the surrounding environment, thereby broadening the scope of cognitive resources. The \textbf{Relevance} Module integrates relevant information from various cognitive tools to produce a comprehensive cognitive output. This synergy aligns with the distributed cognition theory, which posits that cognition is distributed among individuals, tools, and environments. Through this conceptual alignment, our \textbf{QR-CLIP} outperforms previous SOTA methods, achieving an average relative improvement of approximately 10\% and 110\%, respectively. However, the model could struggle with images that contain rich implicit information, and the reasoning model's performance is contingent on the quality and quantity of available knowledge. Insufficient knowledge may also impede the reasoning results. Future work will focus on enhancing reasoning capabilities by refining its architecture and algorithms to handle more complex tasks. Furthermore, incorporating more open-world knowledge will augment the accuracy of reasoning.

\vspace{2.5cm}

\nocite{langley00}

\bibliographystyle{IEEEtran}
\bibliography{IEEE_main}

\newpage

\vspace{11pt}

\vfill

\end{document}